%% file: main.tex
\pdfoutput=1

\documentclass[11pt]{article}

\usepackage[final]{acl}

\usepackage{times}
\usepackage{latexsym}
\usepackage[T1]{fontenc}
\usepackage[utf8]{inputenc}
\usepackage{microtype}
\usepackage{inconsolata}
\usepackage{graphicx}
\usepackage{amssymb}
\usepackage{booktabs}
\usepackage{amsmath}
\usepackage{algorithm}
\usepackage{algpseudocode}
\usepackage{lipsum}
\usepackage{multirow}
\usepackage{cleveref}
\usepackage{placeins}

\input{macros.tex}

\title{Concept Layers: Enhancing Interpretability and \\Intervenability via LLM Conceptualization}

\author{Or Raphael Bidusa, Shaul Markovitch\\
  The Henry and Marilyn Taub Faculty of Computer Science \\
  Technion – Israel Institute of Technology \\
  \texttt{\{bidusa,shaulm\}@cs.technion.ac.il}
 }

\begin{document}
\maketitle
\begin{abstract}


The opaque nature of Large Language Models (LLMs) has led to significant research efforts aimed at enhancing their interpretability, primarily through post-hoc methods. More recent in-hoc approaches, such as Concept Bottleneck Models (CBMs), offer both interpretability and intervenability by incorporating explicit concept representations. However, these methods suffer from key limitations, including reliance on labeled concept datasets and significant architectural modifications that challenges re-integration into existing system pipelines.
In this work, we introduce a new methodology for incorporating interpretability and intervenability into an existing model by integrating Concept Layers (CLs) into its architecture. Our approach projects the model’s internal vector representations into a conceptual, explainable vector space before reconstructing and feeding them back into the model. Furthermore, we eliminate the need for a human-selected concept set by algorithmically searching an ontology for a set of concepts that can be either task-specific or task-agnostic.
We evaluate CLs across multiple tasks, demonstrating that they maintain the original model’s performance and agreement while enabling meaningful interventions. Additionally, we present a proof of concept showcasing an intervenability interface, allowing users to adjust model behavior dynamically, such as mitigating biases during inference. 


\end{abstract}

\section{Introduction}

Large Language Models (LLMs) utilize large amounts of training data, learning rich, high-dimensional embeddings that pass through the network as intricate vector representations
\citep{devlin2019bertpretrainingdeepbidirectional, 
liu2019robertarobustlyoptimizedbert,
raffel2023exploringlimitstransferlearning, vaswani2023attentionneed}. 
While being highly effective, these intricate data representations pose a significant challenge in understanding and explaining the model's reasoning. The "black box" nature has raised concerns about the unchecked deployment of neural networks in areas that demand accountability and transparency such as in healthcare, finance, and legal systems \citep{Golgoon_2024, chen2024surveylargelanguagemodels, mohammadi2025explainabilitypracticesurveyexplainable}.

\emph{Interpretability} is the ability to understand and reason about the model’s decisions. A popular approach for interpretability is post-hoc analysis, which attempts to explain an already-trained model without modifying its computational process \citep{ribeiro2016whyitrustyou, belinkov-glass-2019-analysis, Madsen_2022}. While widely used, post-hoc methods often fall short in capturing the true reasoning process of the model, operating externally, analyzing a model that remains inherently opaque, rather than being an inherent part of the model’s decision-making pipeline \citep{laugel2019dangersposthocinterpretabilityunjustified, Bordt_2022, wei2024revisitingrobustnessposthocinterpretability}.

An alternative in-hoc approach uses \emph{interpretable vector representations}, where each dimension corresponds to a human-understandable concept, embedding interpretability directly into the model’s structure rather than inferring it externally. This paradigm enables both interpretability and \emph{intervenability}, as modifying the interpretable representation during inference allows for potentially mitigating undesirable behaviors. A notable example of this paradigm is the Concept Bottleneck Model (CBM) framework \citep{koh2020conceptbottleneckmodels, chauhan2023interactiveconceptbottleneckmodels, yuksekgonul2023posthocconceptbottleneckmodels,oikarinen2023labelfreeconceptbottleneckmodels, ismail2024concept}, which first maps the input into an interpretable conceptual space, where each dimension represents the semantic relation to a specific concept. The model then uses this conceptual representation to make predictions. While CBMs have been particularly popular in computer vision, only a limited number of works have adapted them to NLP \citep{tan2023interpretingpretrainedlanguagemodels, sun2024conceptbottlenecklargelanguage,ludan2024interpretablebydesigntextunderstandingiteratively}.

Despite their advantages, CBMs suffer from several drawbacks: \emph{Training data constraints} arise as CBMs often require datasets with explicit concept labels  
$(x,C,y)$ to train models to first predict concept representations before making final predictions; \emph{Selecting a set of concepts} is challenging, often requiring domain experts or an external LLM, reintroducing trust issues in black-box models; \emph{Task specificity} limits the applicability of selected concepts to other domains; \emph{Backward compatibility} \& \emph{architectural continuity} pose significant challenges, as LLMs are already integrated into critical applications, and modifying their architecture disrupts existing pipelines, making full adoption impractical.
Some works have addressed these issues, but none have tackled all of them.

In this work, we propose a method of \emph{enhancement} that makes any language model both interpretable and intervenable, rather than building a new model based on on a given language model. Our method integrates Concept Layers (CLs) into the model, enabling conceptual projections and interventions at any layer of the original network. Our approach:

\begin{itemize}
    \item Maintains performance and agreement with the original model.
    \item Does not rely on predefined $(x,C,y)$ datasets for concept mapping and adds no additional learned parameters to the model.
    \item Allows for task-agnostic or task-specific conceptualization, making it widely applicable.
    \item Preserves architectural continuity, ensuring seamless integration with existing systems.
    \item Facilitates automatic selection of concepts through ontology-based search.
\end{itemize}
Our method achieves a structured, hierarchical interpretability framework without disrupting an existing one.

\section{Conceptualizing Language Models}

In this section, we describe our new methodology for enhancing a model’s interpretability and intervenability via conceptualization. Given a model $\fc$ and a concept set $C$, we first show how to construct a \emph{Concept Layer} and integrate it into the model's architecture. In the next section, we introduce a novel, automatic, ontology-based method for generating such a concept set, tailored for the model.

\subsection{Assumptions and Definitions}
Consider a fully trained model 
$ \fc : T \to \mathcal{Y} $, 
parametrized by $ \theta $, that maps text from a textual space $ T $ into a space $ \mathcal{Y} $. 
Define a model slice 
$ \langle \gc, \hc \rangle $
such that 
$ \fc = \hc \circ \gc $
, where 
$ \gc: T \to L $ 
is a prefix of the model, parametrized by 
$ \theta_1 $
, and 
$ \hc: L \to \mathcal{Y} $ 
is a suffix of the model, parametrized by 
$ \theta_2 $. 
The space $L$ of dim $h$ is the model’s internal latent representation, which we aim to project into an interpretable space. 
Denote the given concept set by $C=\{c_1,...,c_n\}$. We assume that we are given a textual semantic representation of each concept $\tau: C \to T$ \cite{simhi2023interpretingembeddingspacesconceptualization}. 
It can be the concept's name, a definition, or a corpus of related texts. The normalized latent representation of a concept $c$  is denoted by $\hat{c}\in L$. Formally, 
$$
\hat{c} \triangleq 
\frac
    {\gc(\tau(c))}
    {\| \gc(\tau(c)) \|}
$$

\begin{figure}[t]
  \includegraphics[width=\columnwidth]{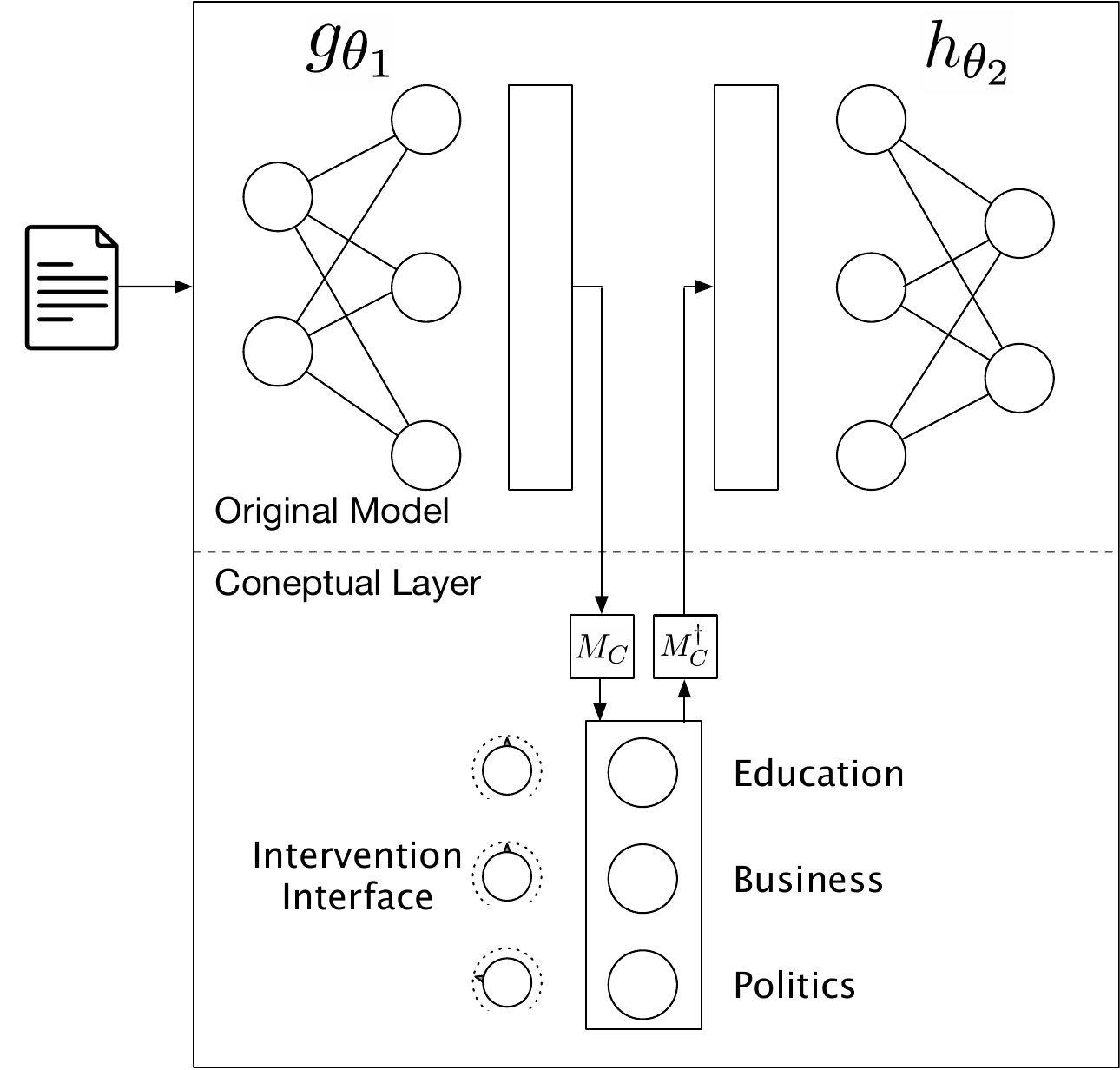}
  \caption{Our methodology, visualized on the task of credit approval. The "Politics" concept can be attenuated in the conceptual space to mitigate bias before passing the vector back to the model.}
  \label{fig:conceptualization_process}
\end{figure}

\subsection{Defining the Concept Layer}
A Concept Layer (CL) is a non-trainable module that integrates into the original model. The CL first projects vectors from the latent space $L$ into an interpretable conceptual space $L_C$, then reconstructs a representation in $L$ before passing it back to $\hc$. $L_C$ is a conceptual space of dimension $n$, where the $i$-th element represents the semantic similarity to the concept $c_i$.
Let $l=\gc(t)$ be the latent representation of some text $t\in T$ in $L$. Formally, CL uses two projections, $\mc :L \to L_C$ and $\mct: L_C\to C$.
$$
\mc(l) \triangleq 
\langle 
\hat{c_1}\cdot l,
...,
\hat{c_n}\cdot l
\rangle
$$

By definition, $\mc(l)$ is a vector of the cosine similarities in the latent space $L$, between the input text and each concept in $C$, factored by the size of $l$. Therefore, in order to get the interpretable representation of a text during inference, all that is left to do is to store $\mc(l)\mathbin{/}\|l\|$.
Note that the projection is a linear operation and therefore $\mc$ can be simply defined as the matrix
$
\mc=
\langle 
\hat{c_1},
...,
\hat{c_n}
\rangle^T
$. 
$\mct$ is the pseudo-inverse of $\mc$, calculated only once upon creating the CL.

As shown in \cref{fig:conceptualization_process} a conceptualization of a model $\fc$ is defined as the process of slicing the model and then integrating a CL in between the slices. This is followed by a short training phase to "weld" the CL to the model by adapting the parameters of $\hc$, as described in the next subsection. The added projections are static and therefore no additional parameters are added to the network. Formally, the conceptualized model of $\fc$ is defined by
$$
\fc^C(t) \triangleq \hcc(\mct\mc\gc(t)), \forall t\in T
$$

\subsection{Welding the CL}

To keep $L_C$ interpretable we want to choose a relatively small number of concepts. $n$ is usually smaller than the hidden dimension $h$, meaning $\mc$ is not a square matrix, and $l$ is projected into a lower-dimensional space. Furthermore, since the concepts themselves might not be independent, their representations in the latent space of the original model may also be correlated, potentially making $\mc$ not well-conditioned. The projection to a conceptual space and back, therefore, limits the expressiveness of the original latent representations by forcing them to align with interpretable concepts, limiting the degrees of freedom of the model. This can be seen as a \emph{structural regularization}, similar to methods like low-rank approximations. \citep{6638949, hu2021loralowrankadaptationlarge}.

Losing too much information, however, can be harmful to the model's performance. A short training phase is required in order to "weld" the CL to the model, adapting $\hc$ to the loss of information. The welding is performed by training $\hc$ for the task of feature-based distillation \citep{romero2015fitnetshintsdeepnets} with regard to the original model. This means that the loss is defined to be the distance between the vectors passing through both models at each stage of the forward pass, rather than just the final representation as in the original distillation method \citep{hinton2015distillingknowledgeneuralnetwork}. This, together with already being close to the original model, results in a fast welding phase on a much smaller dataset than the original model was trained on.

Note that $\gc$ must remain frozen during the welding phase, as training it would change the semantic meaning of the $\mc$ projection. The matrix $\mc$ was constructed by concatenating $\{\hat{c_i}\}_{i=1}^n$ together where each $\hat{c_i}$ was computed using $\gc$ itself. Modifying $\gc$ will prevent $\mc$ from correctly capturing the cosine-similarity between the input text and the concepts.

\subsection{Multi-Layer Conceptualization}

As was shown by post-hoc interpretability methods, different layers in deep neural networks capture different semantic ideas \citep{pmlr-v97-guan19a}. Therefore, in order to enhance the interpretability of the model even more, the process of conceptualization can be repeated on a different layer, on an already conceptualized model. 
Such a process can extract more information about the similarity of an input text to different concepts, as learned by the original model. Furthermore, this process enables intervention at different stages of the forward computation, allowing adjustments to specific conceptual representations as they evolve within the model.

Let $\fc^C$ be an already conceptualize model,
$ \langle \gc', \hc' \rangle $ a slice of $\fc^C$, and $C'$, an additional concept set for the new CL. Note that the new projection matrix, $M_{C'}$ should be calculated with regard to the new prefix $\gc'$ in order to preserve the cosine-similarity semantics of the new CL. An additional important note is that the new slicing should be in a deeper layer in the model than the previous slicing point. Failing to do so will result in disrupting the cosine-similarity semantics of the previous CL in the welding phase.

\section{Concept Set Generation}
We assume that we are given an ontology -- an hierarchical set of human concepts.  Our goal is to select a concept set of size $n$ out of this ontology.
Choosing the right concept set is a crucial step in the conceptualization process. It will define the pivotal ideas by which the input will be interpreted, compared, and projected to. It will determine the types of interventions that could be conducted and affect the output. Since the interpretable vectors are part of the internal architecture, it will also affect the model's performance and expressiveness.

\subsection{Desired Properties}
Given a contextual corpus of texts $\tc$ we want to search for a concept set $C$ that will satisfy the following:
\begin{itemize}
    \item The concepts in $C$ should capture the core ideas within $\tc$, ensuring that they represent the most significant elements of the corpus.
    \item The concepts in $C$ should differentiate between distinct ideas in $\tc$ enabling a clear separation between conceptual regions in the representation space.
    \item The subspace of $L_C$, induced by the projected texts of $\tc$, should be expressive enough to preserve meaningful structure and maintain sufficient variance, ensuring that the conceptual space does not collapse into a limited subspace.
\end{itemize}

\subsection{Task-Specific and Task-Agnostic Models}
The corpus  $\tc$ consists of samples from a text distribution. 
By basing the search on $\tc$, we introduce two distinct conceptualization processes.  If $\tc$ is drawn from a distribution associated with a particular task (e.g., AG News), the resulting conceptualization is classified as task-specific, providing focused interpretability within a particular domain.
Conversely, if $\tc$ is sampled from a generic distribution, preferably the original model’s training set, it will lead to a task-agnostic conceptualization, preserving the model’s versatility while maintaining a more general interpretability.

\subsection{Ontology-Based Search}
 In our context, an ontology is a structured representation of the human knowledge that defines the relation between different concepts\footnote{For the experiments described in the following section we use the English Wikipedia Category Graph.}. Let $G=(C^*,E)$ be our ontology graph, where $C^*$ is a set of concepts and $E$ is the "type-of" relation, meaning that $(c,c')\in E$ if $c'$ is "type of" $c$. Building on the idea of \citet{simhi2023interpretingembeddingspacesconceptualization}, we select a concept set $C$ via a search algorithm over the concept space $C^*$. 
 We denote the set of the successors of a concept $c$ by $Succ(c) \triangleq \{c'\in C^* | (c,c')\in E\}$.

\subsection{Variance-Guided Algorithm}
Our search algorithm maintains a concept set $C_f$, which is returned at the end. The algorithm will also maintain a priority queue of concepts $open$ and a close set $close$ to avoid expanding the same concept twice. It will start with an initial set of concepts as an initial guess, by default the root of the ontology, and in each step will decide which concept out of $open$ should be expanded. The algorithm will also decide which of a concept's successors should be added to $C_f$ and to $open$ itself, to possibly be expanded later. The priority queue will use a variance-based metric called Average Variance Gain (AVG). Let $c$ be a concept. The variance of a corpus $\tc$ with respect to concept $c$, denoted by 
$\mathbf{V}_{T}(c)$, is defined as the variance of the set of projected values:
$$
\mathbf{V}_T(c) \triangleq \operatorname{Var}(
\{ \hat{c} \cdot \gc(t) | \forall t\in \tc\}
)
$$
This measures how well the concept
$c$ spans the semantic variability of $\tc$ in the latent space $L$. A higher variance indicates that $c$ differentiates between diverse meanings within the corpus. Let $s\in Succ(c)$ be a successor of $c$. We will define the Variance Gain (VG) by,
$$
\mathbf{VG}(c,s) \triangleq 
\mathbf{V}_T(s) - \mathbf{V}_T(c)
$$
\input{tables/accuracy_table.tex}

\input{tables/agreement_table.tex}

This measures the additional variance introduced by the successor concept compared to its parent. This measure was influenced by the information gain metric used in algorithms for creating decision trees \cite{Quinlan1986} and serves as a criterion for evaluating how much the addition of a child concept increases the expressiveness of the future $L_C$. We define the Eligible Successors (ES), as the set of successors of $c$ whose Variance Gain exceeds a given threshold $thr$, ensuring they contribute meaningfully to the conceptual space. Formally,
$$
\mathbf{ES}(c,thr) \triangleq 
\{
s\in Succ(c) | \mathbf{VS}(c,s)>thr,  s\notin C_f
\}
$$

This guarantees that only conceptually informative and previously unexplored successors are considered for expansion. Finally, the Average Variance Gain (AVG),
$$
\mathbf{AVG}(c,thr) \triangleq 
\frac{1}{|\mathbf{ES}(c,thr)|}
\sum_{s \in \mathbf{ES}(c, thr)} \mathbf{VG}(c, s)
$$

measures the overall informativeness of expanding a concept by averaging the Variance Gain across its eligible successors. Concepts will be selected from $open$ based on their AVG score until $C_f$ reaches the desired size. If $open$ is exhausted before $C_f$ reaches the target size, the threshold $thr$ is reduced, and open is reinitialized with all concepts currently in $C_f$. This adjustment allows for a broader exploration of the conceptual space by including successors with lower Variance Gain, controlled by a threshold scheduler. We employ a linear scheduler, allowing $thr$ to eventually become negative, enabling non-greedy expansions that may lead to more meaningful expansions in later iterations. The complete pseudocode is provided in Appendix \cref{alg:conceptual_search}.

\section{Experiments}

\input{tables/compatibility_table}
In this section, we evaluate our conceptualization method. First, we assess whether it preserves the original model’s performance, ensuring that enhancement does not degrade accuracy while maintaining agreement with the original predictions.
Second, we verify backward compatibility by testing the enhanced models in the same environment as the original. Finally, we provide a short proof-of-concept evaluation of intervenability.
For our experiments, we used the all-MiniLM-L6-v2 sentence transformer
\footnote{Hugging Face model repository:
\url{https://huggingface.co/sentence-transformers/all-MiniLM-L6-v2}}, 
referred to as "the original model". This model has 6 transformer layers with a hidden size of $384$. We tested our method on three datasets: AG News, Yelp Polarity, and DBpedia-14 \citep{zhang2016characterlevelconvolutionalnetworkstext}. 

Conceptualization was performed at two possible cuts: between the fifth and sixth layers (Single CL) or at both the fourth-to-fifth and fifth-to-sixth layers (Double CL). Concept selection was conducted via an ontology-based search for a set of $100$ concepts. We evaluated eight models in total:

\begin{itemize}
    \item \textbf{Two Task-Agnostic Models}: Single CL and Double CL, trained with task-agnostic conceptualization.
    \item \textbf{Six Task-Specific Models}: Each dataset had two variations: Single CL and Double CL, trained using dataset-specific conceptualization.
\end{itemize}

For the welding process, we used wikitext-103-v1 for task-agnostic models and combined it with the training set for task-specific models. Training was conducted on a single NVIDIA L40S GPU. Task-agnostic models trained faster (around 22 minutes per epoch) as they used only the general corpus, while task-specific models took longer (up to 30 minutes) due to the combined corpus. Models were trained for 15 epochs. Training hyperparameters: batch size = 32, learning rate = 3e-5, optimizer = AdamW, scheduler = linear with 500 warmup steps.

\subsection{Model Recovery}

Enhancing a model’s interpretability is valuable, but it is crucial to ensure that its expressiveness, capabilities, and overall performance remain preserved. To evaluate whether conceptualization affects the model’s effectiveness, we conducted a classification task using our datasets.
For each of the nine models (the original model plus eight conceptualized variants), we trained a separate MLP classification head on the training set. The MLP was trained until convergence, monitored using a validation set.

\subsubsection{Raw Performance}

To assess performance, we evaluated each classifier on the test set, measuring accuracy, weighted F1-score, and loss. The accuracy results for all models are presented in \cref{tab:accuracy_results}, while weighted F1-scores and losses are included in the appendix (\cref{tab:f1_weighted_results,tab:loss_results}).

The results indicate that conceptualization preserves model performance, with conceptualized models performing on par with or slightly better than the original model in most cases. Notably, the best-performing model for each dataset was \emph{a conceptualized variant} rather than the original.

\subsubsection{Agreement}
Maintaining performance is essential, but agreement with the original model is equally critical. Two models may both achieve 90\% accuracy, yet still disagree on 20\% of the predictions if their errors occur on different samples. High agreement ensures that the conceptualized model retains behavioral consistency with the original model.

The agreement rates between the conceptualized and original models across datasets are reported in \cref{tab:agreement_results}. The results demonstrate a high level of agreement, confirming that conceptualization does not introduce drastic behavioral changes.

\subsection{Backward Compatibility}

Language models are deeply integrated into real-world applications, making backward compatibility a critical requirement. Enhancements should seamlessly integrate without disrupting existing components. In addition to maintaining the same output dimensionality, an enhanced model must ensure its outputs remain compatible with downstream modules.

A good indicator of compatibility is whether a classifier head trained \emph{on the original model} remains effective when transferred to a conceptualized model without retraining. To evaluate this, we applied an MLP classifier head, trained on the original model, to each conceptualized variant and measured accuracy and agreement (\cref{tab:backward_compatibility_results}). The corresponding F1-weighted scores and loss values are provided in the appendix 
(\cref{tab:compatibility_f1_results}, \cref{tab:compatibility_loss_results}).
While a slight, expected drop in accuracy was observed, agreement with the original model remained consistently high. Most notably, the task-specific conceptualized models outperformed task-agnostic ones in both accuracy and agreement.

\subsection{Interpretability and Intervenability}

\subsubsection{Interpretability}

 The conceptual vectors in $L_c$ are inherently interpretable as they represent cosine similarities between the input representation and each concept. If the original model learned rich semantic representations, the projection will reflect meaningful relationships. Since $\gc$ remains unchanged during the welding phase, interpretability can be extracted by dividing $\mc(l)$ by $\|l\|$ and then sorting these values and selecting the top $k$ concepts provides an interpretable explanation.

Therefore, a key aspect of interpretability is the concept set itself—and by extension, the method used to select it. Our variance-based heuristic determines which concepts are included in the projection space. While a human study is beyond this work’s scope, we provide a list of selected concepts in the appendix (\cref{tab:ag4,tab:yelp4,tab:db4}) for examination. In Yelp Polarity, the retrieved concepts align with the dataset’s domain—food, museums, arts, pubs, and nightlife—with minimal noise. AG News shows similar alignment. In contrast, DBpedia, which classifies Wikipedia-derived categories, mirrors the Wikipedia category graph, making it unsuitable as an independent interpretability benchmark.

\input{tables/yelp_table_short}

\subsubsection{Interventability}
Intervenability refers to modifying the model’s decision-making by adjusting its conceptual representation. If each vector element corresponds to a distinct aspect, modifying it should produce a predictable, aspect-specific change in behavior. \citet{chauhan2023interactiveconceptbottleneckmodels} demonstrated this by enabling users to query and adjust individual concepts.

We provide a short proof of concept demonstrating model intervenability. We show how modifying conceptual activations influences model predictions and analyze specific cases. 

Our intervention interface is straightforward: it takes a list of pre-selected concepts and, during inference, attenuates the corresponding vector elements by multiplying them with a discount factor. We test this interface on a task-specific Concept Layer trained on the Yelp Polarity dataset.

Imagine a scenario where a Yelp Polarity-based classifier recommends attractions to a user. Suppose the user is unconcerned with cost and does not want price-related biases to affect recommendations. Since the dataset lacks explicit categories for why an attraction is classified as negative, we cannot directly filter results based on price. Instead, we use our intervention mechanism to reduce the influence of the "Economy" concept, ensuring that overpriced attractions are still considered in recommendations.
\Cref{tab:yelp_short} presents examples of reviews that were originally classified as negative (true negatives) but, after intervention, were reclassified as positive, demonstrating the model's ability to adjust predictions in a controlled, concept-driven manner.

\section{Related Work}

\subsection{Post-hoc Conceptualization of Embedding Spaces}

\citet{simhi2023interpretingembeddingspacesconceptualization} proposed a post-hoc method for interpreting model embeddings by mapping them to a concept space, unlike our approach, which integrates conceptualization into the model. Their method introduces an automatic concept selection process using ontology-based search, similar in spirit to ours but with a different selection criterion. Instead of relying on learning from predefined datasets of concepts scores, they map model representations through dot product, avoiding the need for additional learned parameters.

\subsection{Concept Bottleneck Models (CBMs)}

The Concept Bottleneck Model (CBM) framework was introduced by \citet{koh2020conceptbottleneckmodels}, proposing a structured approach where models first predict human-interpretable concepts before making final decisions. This allows for transparency and direct intervention at the concept level. CBMs have been widely explored, mainly in vision tasks. However, they require explicitly labeled concept datasets $(x, C, y)$ and are inherently task-specific. Moreover, they define a new end-to-end model architecture rather than working with existing models, requiring full retraining.

\subsection{Extending CBMs: Interactive and Label-Free Approaches}

Interactive CBMs (ICBMs) \citep{chauhan2023interactiveconceptbottleneckmodels} extend the CBM framework by introducing human feedback at inference time, allowing users to adjust concept activations before final predictions. This system enhances intervenability, as it enables real-time corrections to improve decision-making.

Label-Free CBMs (LF-CBMs) \citep{oikarinen2023labelfreeconceptbottleneckmodels} focus on automating concept selection and labeling. Instead of requiring predefined concepts, LF-CBMs query an external LLM to generate concepts dynamically. Concept scores are then inferred using CLIP-based similarity. This approach removes the dependency on manually labeled datasets while still following the last-layer bottleneck structure.

\subsection{Concept Bottlenecks in NLP}

Recent works have explored adapting CBMs to natural language processing. Concept Bottleneck Large Language Models (CB-LLMs) \citet{sun2024conceptbottlenecklargelanguage} introduced concept bottlenecks into LLMs, demonstrating their applicability in both text classification and text generation tasks. By enforcing conceptual constraints on latent representations, CB-LLMs enable interpretability while allowing for structured reasoning within LLM architectures. However, they rely on an external LLM for concept generation, remain task-specific, and following the last-layer bottleneck structure.

\citet{ludan2024interpretablebydesigntextunderstandingiteratively} introduced Text Bottleneck Models (TBMs), an interpretable text classification framework where a linear predictor is trained on concept labels generated by GPT-4. This approach relies on an external LLM to define and label the concept space, This approach depends entirely on GPT-4 for concept definition and labeling, making it reliant on external querying rather than internalizing a conceptual representation within the model.

Another approach, C3M \cite{tan2023interpretingpretrainedlanguagemodels}, merges human-annotated concepts with concepts generated and labeled by ChatGPT to build a CBM on top of GPT-2 and BERT. By integrating human-defined and generated concepts, C3M provides a flexible way to incorporate structured reasoning in NLP tasks. However, it still requires predefined concept labels and relies on external models for generating part of the concept space.

\subsection{Maintaining Model Structure Through Conceptual Mapping}

Recent works have explored integrating concepts into existing models without enforcing a strict bottleneck while preserving their original structure. The framework suggested by \citet{laguna2024conceptbottleneckmodelsmake} enables modifying model behavior through concept-based interventions without altering the underlying model. However, this framework still requires labeled $(x, C, y)$ validation set for probing.

AnyCBMs \citep{dominici2024anycbmsturnblackbox} propose a post-hoc method to transform any pretrained model into a CBM-like system without requiring full retraining, By mapping internal model embeddings into a conceptual space. However, AnyCBMs rely on a validation set and use concepts generated by GPT-3, reintroducing dependencies on external models.

Both approaches preserve the structure of existing models, mapping internal representations into a conceptual space instead of enforcing an explicit concept bottleneck. However, they still require external supervision through labeled validation sets or predefined concept sets.

\section{Conclusions}
In this paper, we presented a novel methodology for enhancing a given LLM by incorporating conceptual layers into its architecture. We demonstrated that our approach introduces interpretability and intervenability without degrading the original model's performance.

We believe that our method will enable the development of techniques that leverage our intervention interface for understanding, debugging, and detecting biases in existing models. In future work, we plan to extend our experiments to more resource-intensive generative models.

\section{Limitations}
The welding phase introduces a necessary adaptation step where the model aligns with the conceptualized representation, requiring a short training process. This phase relies on distillation from the original model, which demands either direct access to its latent representations during training or precomputing them in advance. Both approaches can be challenging depending on system constraints, making this non-trivial step the primary computational cost of our method.

\section{Ethical Considerations}

The new methodology presented here has the potential to positively impact a wide range of ethical issues. For instance, our intervention interface can enable users of job application filtering systems to minimize the influence of political factors in decision-making by discounting politics-related concepts.

\bibliography{custom}

\clearpage

\appendix

\section*{Appendix}
\input{conceptual_search}

\clearpage


\input{tables/f1_table.tex}

\input{tables/loss_table.tex}

\input{tables/f1_comp_table}


\input{tables/loss_comp_table}

\clearpage


\input{concept_lists/ag4.tex}

\input{concept_lists/yelp4.tex}

\input{concept_lists/db4.tex}

\end{document}

%% file: macros.tex
\newcommand{\fc}{f_{\theta}}

\newcommand{\gc}{g_{\theta_1}}
\newcommand{\hc}{h_{\theta_2}} 
\newcommand{\hcc}{h_{\theta_2'}} 
\newcommand{\mc}{M_{C}} 
\newcommand{\mct}{M^{\dagger}_{C}} 
\newcommand{\tc}{T_{\text{context}}} 

%% file: tables/accuracy_table.tex
\begin{table*}[h]
    \centering
    \begin{tabular}{|l|l|c|c|c|}
        \hline
        \multicolumn{2}{|c|}{} & \textbf{AGnews} & \textbf{Yelp} & \textbf{DBpedia} \\
        \hline
        \multicolumn{2}{|l|}{\textbf{Original}} & 91.63 & 90.78 & 98.25 \\
        \hline
        \multirow{2}{*}{\textbf{Task-Agnostic}} & Single CL & 91.78 & 90.44 & \textbf{98.39} \\
        & Double CL & 91.75 & 90.54 & 98.32 \\
        \hline
        \multirow{2}{*}{\textbf{Task-Specific}} & Single CL & \textbf{91.86} & 90.84 & 98.32 \\
        & Double CL & 91.78 & \textbf{91.11} & 98.32 \\
        \hline
    \end{tabular}
    \caption{Classification accuracy across different models and datasets.}
    \label{tab:accuracy_results}
\end{table*}

%% file: tables/agreement_table.tex
\begin{table*}[h]
    \centering
    \begin{tabular}{|l|l|c|c|c|}
        \hline
        \multicolumn{2}{|c|}{} & \textbf{AGnews} & \textbf{Yelp} & \textbf{DBpedia} \\
        \hline
        \multirow{2}{*}{\textbf{Task-Agnostic}} & Single CL & 95.57 & 94.28 & \textbf{98.86} \\
        & Double CL & 95.42 & 93.22 & 98.68 \\
        \hline
        \multirow{2}{*}{\textbf{Task-Specific}} & Single CL & 95.89 & \textbf{94.43} & 98.80 \\
        & Double CL & \textbf{96.45} & 94.04 & 98.70 \\
        \hline
    \end{tabular}
    \caption{Agreement with the original model.}
    \label{tab:agreement_results}
\end{table*}

%% file: tables/compatibility_table.tex
\begin{table*}[h]
    \centering
    \begin{tabular}{|l|l|c|c|c|}
        \hline
        \multicolumn{2}{|c|}{} & \textbf{AGnews} & \textbf{Yelp} & \textbf{DBpedia} \\
        \hline
        \multirow{2}{*}{\textbf{Task-Agnostic}} & Single CL & 90.91 / 95.92 & 87.66 / 91.51 & 98.06 / 98.91 \\
        & Double CL & 90.79 / 95.43 & 87.69 / 90.97 & 97.87 / 98.57 \\
        \hline
        \multirow{2}{*}{\textbf{Task-Specific}} & Single CL & 90.92 / \underline{96.14} & 89.50 / \underline{94.41} & \textbf{98.09} / \underline{98.96} \\
        & Double CL & \textbf{90.96} / 95.91 & \textbf{89.77} / 94.01 & 97.95 / 98.74 \\
        \hline
    \end{tabular}
    \caption{Backward compatibility results: The left value is accuracy and the right value is agreement with the original model. Best accuracy is in bold, and best agreement is underlined.}
    \label{tab:backward_compatibility_results}
\end{table*}

%% file: tables/yelp_table_short.tex
\begin{table*}[h]
    \centering
    \caption{Comparison of Positive Aspects and Price Complaints}
    \begin{tabular}{p{7cm} | p{7cm}}
        \toprule
        \textbf{Positive / Natural Aspects of the Attraction} & \textbf{Price-Related Complaints} \\
        \midrule
        \textquotedblleft The portions that came with my meal were just the way I like them... The flavor of the oysters and shrimp was above satisfactory.\textquotedblright & \textquotedblleft The prices were on the high side of expensive... I would have kicked myself if I had to pay full price for the experience received.\textquotedblright \\
        \hline
        \textquotedblleft "I have to say that in general, the food wasn't bad, but it wasn't anything special..."\textquotedblright & \textquotedblleft "...but it was very expensive."\textquotedblright \\
        \hline
        \textquotedblleft The waiter was helpful and courteous. Good Caesar salad.\textquotedblright & \textquotedblleft 7.25 for a margarita and 11.00 for a 6 shrimp cocktail??... I dropped 60.00 on dinner and had a burger, fries, and water.\textquotedblright \\
        \hline
        
    \end{tabular}
    \label{tab:yelp_short}
\end{table*}

%% file: conceptual_search.tex
\begin{algorithm}[h]
\caption{Conceptual Search Algorithm}
\label{alg:conceptual_search}
\textbf{Input:} Initial concept set \( C_{init} \), threshold scheduler \( thr\_scheduler \), ontology graph \( G = (C^*, E) \), target\_size for \( |C_f| \) \\
\textbf{Output:} Expanded concept set \( C_f \)

\begin{algorithmic}[1]
\State \( C_f \gets C_{init} \)
\While{\( |C_f| < \text{target\_size} \)}
    \State \( thr \gets thr\_scheduler.\text{next}() \) 
    \State \( open \gets \emptyset \), \( close \gets \emptyset \) 
    
    \For{\( c \in C_f \)}
        \State \( S_c \gets \mathbf{ES}(c, thr) \)
        \State \( \text{score} \gets \mathbf{AVG}(c, thr) \)
        \State \( open.\text{push}((c, S_c), \text{score}) \)
    \EndFor
    
    \While{\( open \neq \emptyset \)}
        \State \( (c, successors) \gets open.\text{pop}() \) 
        \State \( C_f \gets C_f \cup \text{successors} \)
        \State \( close \gets close \cup \{c\} \)

        \For{\( s \in \text{successors} \)}
            \If{\( s \notin close \) \textbf{and} \( s \notin open \)}
                \State \( S_s \gets \mathbf{ES}(s, thr) \)
                \State \( \text{score} \gets \mathbf{AVG}(s, thr) \)
                \State \( open.\text{push}((s, S_s), \text{score}) \)
            \EndIf
        \EndFor
    \EndWhile
\EndWhile
\If{$|C_f| > \text{target\_size}$}
    \State Remove the last $(|C_f| - \text{target\_size})$ added concepts from $C_f$
\EndIf
\State \Return \( C_f \)
\end{algorithmic}
\end{algorithm}

%% file: tables/f1_table.tex
\begin{table*}[h]
    \centering
    \begin{tabular}{|l|l|c|c|c|}
        \hline
        \multicolumn{2}{|c|}{} & \textbf{AGnews} & \textbf{Yelp} & \textbf{DBpedia} \\
        \hline
        \multicolumn{2}{|l|}{\textbf{Original}} & 91.63 & 90.78 & 98.25 \\
        \hline
        \multirow{2}{*}{\textbf{Task-Agnostic}} & Single CL & 91.76 & 90.44 & \textbf{98.39} \\
        & Double CL & 91.74 & 90.54 & 98.32 \\
        \hline
        \multirow{2}{*}{\textbf{Task-Specific}} & Single CL & \textbf{91.85} & 90.84 & 98.32 \\
        & Double CL & 91.77 & \textbf{91.11} & 98.32 \\
        \hline
    \end{tabular}
    \caption{F1 weighted scores across different models and datasets. Values are multiplied by 100.}
    \label{tab:f1_weighted_results}
\end{table*}

%% file: tables/loss_table.tex
\begin{table*}[h]
    \centering
    \begin{tabular}{|l|l|c|c|c|}
        \hline
        \multicolumn{2}{|c|}{} & \textbf{AGnews} & \textbf{Yelp} & \textbf{DBpedia} \\
        \hline
        \multicolumn{2}{|l|}{\textbf{Original}} & 0.2461 & 0.2235 & 0.0622 \\
        \hline
        \multirow{2}{*}{\textbf{Task-Agnostic}} & Single CL & 0.2495 & 0.2283 & 0.0560 \\
        & Double CL & 0.2444 & 0.2266 & 0.0576 \\
        \hline
        \multirow{2}{*}{\textbf{Task-Specific}} & Single CL & 0.2467 & 0.2252 & 0.0578 \\
        & Double CL & 0.2429 & 0.2170 & 0.0592 \\
        \hline
    \end{tabular}
    \caption{Loss values across different models and datasets.}
    \label{tab:loss_results}
\end{table*}

%% file: tables/f1_comp_table.tex
\begin{table*}[h]
    \centering
    \begin{tabular}{|l|l|c|c|c|}
        \hline
        \multicolumn{2}{|c|}{} & \textbf{AGnews} & \textbf{Yelp} & \textbf{DBpedia} \\
        \hline
        \multirow{2}{*}{\textbf{Task Agnostic}} & Single CL & 90.91 & 87.66 & 98.05 \\
        & Double CL & 90.77 & 87.69 & 97.86 \\
        \hline
        \multirow{2}{*}{\textbf{Task Specific}} & Single CL & 90.91 & 89.50 & 98.09 \\
        & Double CL & 90.96 & 89.77 & 97.94 \\
        \hline
    \end{tabular}
    \caption{Compatibility F1-score across different models and datasets. Values are multiplied by 100}
    \label{tab:compatibility_f1_results}
\end{table*}

%% file: tables/loss_comp_table.tex
\begin{table*}[h]
    \centering
    \begin{tabular}{|l|l|c|c|c|}
        \hline
        \multicolumn{2}{|c|}{} & \textbf{AGnews} & \textbf{Yelp} & \textbf{DBpedia} \\
        \hline
        \multirow{2}{*}{\textbf{Task Agnostic}} & Single CL & 0.2741 & 0.2941 & 0.0717 \\
        & Double CL & 0.2754 & 0.2933 & 0.0762 \\
        \hline
        \multirow{2}{*}{\textbf{Task Specific}} & Single CL & 0.2757 & 0.2563 & 0.0710 \\
        & Double CL & 0.2712 & 0.2476 & 0.0748 \\
        \hline
    \end{tabular}
    \caption{Compatibility loss values across different models and datasets.}
    \label{tab:compatibility_loss_results}
\end{table*}

%% file: concept_lists/ag4.tex
\begin{table*}[h]
\centering
\begin{minipage}{0.48\textwidth}
\centering
\begin{tabular}{|l|}
\hline
World \\
Countries \\
Former countries \\
Countries by international organization \\
Countries in fiction \\
Turkic states \\
Global studies \\
Global studies research \\
Global culture \\
Cultural globalization \\
Global citizenship \\
Continents \\
Politics by continent \\
Continental fragments \\
Antarctica \\
Fictional continents \\
International organizations \\
International law organizations \\
Global governance \\
Global politics \\
Politics \\
Political culture \\
Fascism \\
Liberalism \\
Democratic socialism \\
Social liberalism \\
Communism \\
Political communication \\
Political activism \\
Sports \\
Sports venues \\
Sports venue logos \\
Sports complexes \\
Disasters in sports venues \\
Sports venue architects \\
Sports venue managers \\
Sports by century \\
Gaelic games by century \\
Sport by year \\
Sports events \\
Sporting events by country \\
Recurring sporting events \\
Defunct sporting events \\
Sports by year \\
Sports administration \\
Cricket administration \\
Chess patrons \\
Military sport \\
Military sports teams \\
Military sports competitions \\
\hline
\end{tabular}
\end{minipage}
\hfill 
\begin{minipage}{0.48\textwidth}
\centering
\begin{tabular}{|l|}
\hline
Military sports clubs \\
Military association football \\
Sports instruction \\
Sports techniques \\
Banned sports tactics \\
Politics and sports \\
Sports journalism \\
Sports festivals \\
Ancient Greek athletic festivals \\
Equestrian festivals \\
Sports strategy \\
Ice hockey strategy \\
Sports seasons \\
Sports team seasons \\
Sports accomplishments \\
Business \\
Business planning \\
Business process \\
Business culture \\
Industries \\
Business ownership \\
International business \\
Business economics \\
Science \\
Fringe science \\
Scientific phenomena \\
Scientific folklore \\
Scientific classification \\
Scientific disciplines \\
Scientific organizations \\
Technology \\
Propaganda \\
Sports plays \\
Scientific speculation \\
Technology evangelism \\
Diplomacy \\
Sailing festivals \\
Quotations from science \\
Socialism \\
Sports by decade \\
Political corruption \\
Electoral fraud \\
Voter suppression \\
Ethically disputed political practices \\
Governance \\
Bribery \\
Political institutions \\
Sport by decade \\
Kite festivals \\
Political scandals \\
\hline
\end{tabular}
\end{minipage}

\caption{Concept Set, AG news, Single CL}
\label{tab:ag4}
\end{table*}

%% file: concept_lists/yelp4.tex
\begin{table*}[h]
\centering
\begin{minipage}{0.48\textwidth}
\centering
\begin{tabular}{|l|}
\hline
Sugar museums \\
Main topic classifications \\
Humanities \\
Archaeology \\
Archaeology images \\
History \\
Art history \\
Art history books \\
Islamic art \\
Christian art \\
Ancient artists \\
Entertainment \\
Comedy \\
Comedy venues \\
Comedy genres \\
Amusement parks \\
Miniature parks \\
Theme parks \\
Theatre \\
Theatre awards \\
Theatre festivals \\
Theatres \\
Entertainment by continent \\
Variety shows \\
Talent shows \\
Vaudeville \\
Nightlife \\
Pubs \\
Cabaret \\
Nightclubs \\
Entertainment lists \\
Industry \\
Industrial tourism \\
Industry by country \\
Geography \\
Places \\
Landscape \\
Language \\
Languages \\
World \\
Continents \\
Antarctica \\
Food and drink \\
Dairy \\
Dairy dishes \\
Food and drink museums \\
Agriculture museums \\
Salt museums \\
Chocolate museums \\
Potato museums \\
\hline
\end{tabular}
\end{minipage}
\hfill 
\begin{minipage}{0.48\textwidth}
\centering
\begin{tabular}{|l|}
\hline
Cuisine \\
Vegetarian cuisine \\
Economy \\
Trade \\
The arts \\
Arts awards \\
Arts venues \\
Arts districts \\
LGBT arts \\
Arts by location \\
Arts by culture \\
Jewish art \\
Entertainment by city \\
Entertainment in Chennai \\
Archaeologists \\
Western art \\
Style \\
Religion \\
Gambling \\
Ethnic religion \\
Gambling games \\
National churches \\
Germanic religion \\
Skateparks \\
History images \\
Dairy by country \\
Landscape architecture \\
Comedy tours \\
Archaeology publications \\
Food activism \\
Jewish comedy and humor \\
Dance \\
Dance venues \\
Dance magazines \\
Dance companies \\
Dance by continent \\
Dances \\
Dance equipment \\
Dance organizations \\
Museology \\
Museum design \\
Museum publications \\
Government \\
Veto \\
Entertainment halls of fame \\
Ministries \\
Ministerial offices \\
Government by region \\
Toy halls of fame \\
Religion and geography \\
\hline
\end{tabular}
\end{minipage}

\caption{Concept Set, Yelp Polarity, Single CL}
\label{tab:yelp4}
\end{table*}

%% file: concept_lists/db4.tex
\begin{table*}[h]
\centering
\begin{minipage}{0.48\textwidth}
\centering
\begin{tabular}{|l|}
\hline
Main topic classifications \\
Religion \\
Ethnic religion \\
Ancient Semitic religions \\
Germanic religion \\
Ancient Greek religion \\
Religious identity \\
Religious occupations \\
Religion by period \\
Religion by decade \\
Organizations \\
Organizations awarded Nobel Peace Prizes \\
History of organizations \\
Categories by organization \\
Proposed organizations \\
Government \\
Veto \\
Ministries \\
Governmental studies academics \\
Government research \\
Government corporations \\
Politics \\
Politics by period \\
Governance \\
Political activism \\
Political congresses \\
Activist publications \\
Politics awards \\
Humanities \\
Medical humanities \\
Archaeology \\
Art history \\
Islamic art \\
Art history journals \\
Ancient artists \\
Art history books \\
Christian art \\
Digital humanities \\
Humanities occupations \\
Humor research \\
Humor researchers \\
Fiction \\
Fiction writers \\
Fiction awards \\
Fiction anthologies \\
Linguistics \\
Humanities organizations \\
Humanities conferences \\
Language \\
Languages \\
\hline
\end{tabular}
\end{minipage}
\hfill 
\begin{minipage}{0.48\textwidth}
\centering
\begin{tabular}{|l|}
\hline
Ministerial offices \\
Medical sociology \\
Turkish ministerial offices \\
Art historians \\
Geography \\
Landscape \\
Landscape architecture \\
Geography terminology \\
Geographical technology \\
Political communication \\
Humanities education \\
Language education \\
Landscape ecology \\
Register offices \\
Caretaker governments \\
Cultural education \\
Open government \\
Islamic religious occupations \\
E-government \\
Ancient Celtic religion \\
Categories by religion \\
Dynasties by religion \\
Inscriptions by religion \\
Manuscripts by religion \\
E-democracy \\
Political history \\
Imperialism \\
Fiction forms \\
Political historians \\
Medical anthropology \\
Language software \\
Electoral history \\
Architectural education \\
Propaganda \\
Organizational studies \\
Shinto religious occupations \\
Organizational culture \\
Archaeology publications \\
Politicides \\
Political titles \\
Vice presidents \\
Buddhist religious occupations \\
Presidents \\
Industry \\
Civil services \\
Industrial archaeology \\
Appointments \\
Industrial tourism \\
Industrial history \\
History of taxation \\
\hline
\end{tabular}
\end{minipage}

\caption{Concept Set, DBpedia, Single CL}
\label{tab:db4}
\end{table*}